\documentclass[letterpaper, 10 pt, conference]{ieeeconf}  

\IEEEoverridecommandlockouts                              

\makeatletter
\let\NAT@parse\undefined
\makeatother
\usepackage[numbers,sort&compress]{natbib}

\usepackage[pdftex]{graphicx}
\usepackage{amsmath}
\usepackage{amssymb}  
\usepackage{subfigure}
\usepackage{multirow}
\usepackage{array,booktabs}
\usepackage{diagbox}
\usepackage{balance}
\usepackage{xspace}
\usepackage{algorithm}
\usepackage{algpseudocode}
\usepackage{adjustbox}
\usepackage{romannum}
\usepackage{subfloat}

 \usepackage[final]{hyperref}
 \hypersetup{
     colorlinks=true,
     linkcolor=blue,
     filecolor=magenta,
     urlcolor=blue,
     citecolor=black
 }
\usepackage[labelfont=bf, font=small]{caption}
\usepackage{./rpm_packages/rpm_SIunits}
\usepackage{./rpm_packages/rpm_acronyms}
\usepackage{./rpm_packages/rpm_math}
\usepackage{./rpm_packages/rpm_misc}

\usepackage{soul,color}
\usepackage{lipsum}

\usepackage{xcolor}

\usepackage{tikz} 
\usepackage{subcaption}
\usepackage{caption}
\title{\LARGE \bf The City that Never Settles: Simulation-based LiDAR Dataset for Long-Term Place Recognition Under Extreme Structural Changes}     

\author{Hyunho Song${}^{1}$, Dongjae Lee${}^{2}$, Seunghun Oh${}^{2}$, Minwoo Jung${}^{2}$, and Ayoung Kim${}^{2*}$
\thanks{$^\dagger$This work was supported by the National Research Foundation of Korea(NRF) grant funded by the Korea government(MSIT) (No. RS-2023-00241758)}
\thanks{$^{1}$H. Song is with the Dept. of Future Automotive Mobility, SNU, Seoul, S. Korea {\tt\small hun1021405@snu.ac.kr}}%
\thanks{$^{2}$D.Lee, S.Oh, M. Jung and A. Kim are with the Dept. of Mechanical Engineering, SNU, Seoul, S. Korea {\tt\small [pur22, alvin0808, moonshot, ayoungk]@snu.ac.kr}}%
}
\begin{document}

\makeatletter
  \let\@oldmaketitle\@maketitle
  \renewcommand{\@maketitle}{\@oldmaketitle
  \bigskip
  \centering
    \includegraphics[trim= 0cm 0cm 0cm 0cm, clip,width=0.88\textwidth]{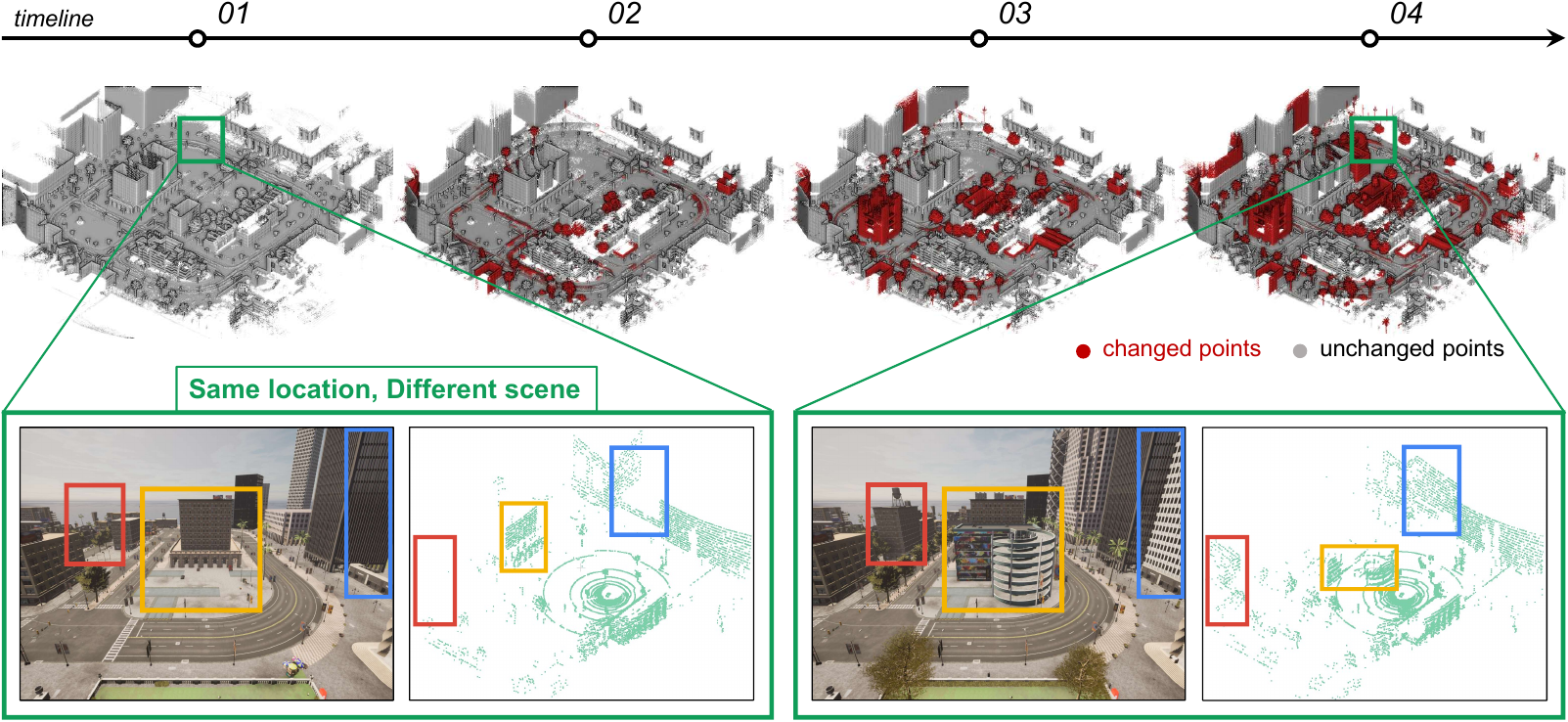}
    \captionof{figure}{
        \textbf{Structural changes over time in }\texttt{Metropolis}\textbf{.} The upper part of the image shows how the city evolves over time through point cloud maps, where gray points represent unchanged areas and red points indicate structural changes. The lower part of the image illustrates two scenes captured at the same location, demonstrating how buildings are constructed over time, resulting in different observations.
    }
    \label{fig:main}
    \vspace{-4mm}
  }
\makeatother

\maketitle
\thispagestyle{empty}
\pagestyle{empty}

\begin{abstract}

Large-scale construction and demolition significantly challenge long-term place recognition (PR) by drastically reshaping urban and suburban environments. Existing datasets predominantly reflect limited or indoor-focused changes, failing to adequately represent extensive outdoor transformations. To bridge this gap, we introduce \ac{CNS} dataset, a simulation-based dataset created using the CARLA simulator, capturing major structural changes—such as building construction and demolition—across diverse maps and sequences. Additionally, we propose TCR$_{\text{sym}}$, a symmetric version of the original TCR metric, enabling consistent measurement of structural changes irrespective of source-target ordering. Quantitative comparisons demonstrate that \ac{CNS} encompasses more extensive transformations than current real-world benchmarks. Evaluations of state-of-the-art LiDAR-based PR methods on \ac{CNS} reveal substantial performance degradation, underscoring the need for robust algorithms capable of handling significant environmental changes. Our dataset is available at 
\href{https://github.com/Hyunho111/CNS\_dataset}{https://github.com/Hyunho111/CNS\_dataset}.

\end{abstract}
\addtocounter{figure}{-1}
\section{Introduction}
\label{sec:intro}

Outdoor environments are dynamic and constantly evolving, with construction activities reshaping not only building interiors but also the entire cityscape. In these continuously changing environments, relying on static maps for localization becomes increasingly difficult. While the \ac{NSS} benchmark \cite{sun2025standsstillspatiotemporalbenchmark} addressed extreme structural changes in construction sites, it focused primarily on interior modifications and did not capture large-scale transformations at the urban and suburban scales. Moreover, acquiring real-world data that reflects such massive evolutionary changes over time is impractical due to the long observation periods and logistical constraints.

To overcome these challenges, we propose the \ac{CNS} dataset, which simulates large-scale structural changes in both urban and suburban environments using the CARLA simulator \cite{dosovitskiy2017carla}. As shown in \figref{fig:main}, we simulate large-scale transformations such as construction and demolition across different cityscapes, providing a new benchmark for testing long-term place recognition (PR). Additionally, we introduce the TCR$_{\text{sym}}$ metric, a modification of the TCR from the NSS benchmark \cite{sun2025standsstillspatiotemporalbenchmark}, to quantitatively compare the rate of change between real-world and our dataset. This enables an objective comparison of the extent of structural variations, highlighting the larger-scale transformations captured by our approach.

Our key contributions are as follows:
\begin{enumerate}
    \item We provides simulation-based dataset which systematically introduces massive construction and demolition across multiple maps and sequences, using CARLA to emulate real-world transformations at scale.

    \item We propose a new metric, TCR$_{\text{sym}}$, ensuring symmetric measurements of structural changes. Comparing real-world datasets with ours demonstrates far larger transformations, reinforcing the dataset’s value for long-term PR.

    \item We evaluate state-of-the-art LiDAR-based PR methods on our dataset and reveal substantial performance drops under these extreme changes, highlighting the need for more robust, adaptable algorithms.
\end{enumerate}

\section{Related Works}
\label{sec:related_work}

Numerous datasets containing temporal and structural variations have been developed to support long-term LiDAR-based PR. The NCLT dataset \cite{carlevaris2016university}, which spanned 27 sessions over a year, captured long-term dynamics including major construction projects. Similarly, the Oxford Robotcar dataset \cite{maddern20171} extended over a year and a half, reflecting structural changes caused by city roadworks, while the Boreas dataset \cite{burnett2023boreas} primarily tracked seasonal variations but also included some construction-related changes. The Mulran \cite{kim2020mulran} and the HeLiPR dataset \cite{jung2024helipr}, collected at the same location with a four-year gap between sessions, revealed notable structural shifts. Although these real-world datasets provide valuable insights, the modifications they depict remain relatively minor, with permanent structures dominating the scene.

Meanwhile, the NSS benchmark \cite{sun2025standsstillspatiotemporalbenchmark} does capture drastic spatiotemporal changes in construction sites. However, it was originally focused on point registration and largely confined to indoor spaces, leaving city-scale outdoor transformations unaddressed. As a result, the existing datasets—both real-world and \ac{NSS}—offer only limited challenges for PR algorithms when faced with massive structural changes.

The \ac{CNS} dataset is specifically designed to address the challenges posed by extreme structural changes at the urban and suburban scale. By leveraging the CARLA simulator \cite{dosovitskiy2017carla}, we simulate large-scale transformations, including construction and demolition processes. This provides a more rigorous evaluation setting for LiDAR-based PR methods, enabling the development of robust algorithms capable of handling massive environmental transformations.
\section{The CNS Dataset}
\label{sec:method}

In this section, we present the \textbf{City that Never
 Settles} (\textbf{CNS}) dataset, which aims to facilitate the investigation of long-term PR in highly dynamic environments. By leveraging the CARLA simulator, we are able to emulate significant structural changes—such as construction and demolition—across multiple sequences of the same map. The data was acquired using an NVIDIA GeForce RTX 4070 GPU and a 13th Gen Intel Core i9-13900HX x 32 CPU.

\begin{figure}[!t]
    \centering
    \includegraphics[trim= 0cm 1.5cm 0cm 0cm, clip,width=0.7\linewidth]{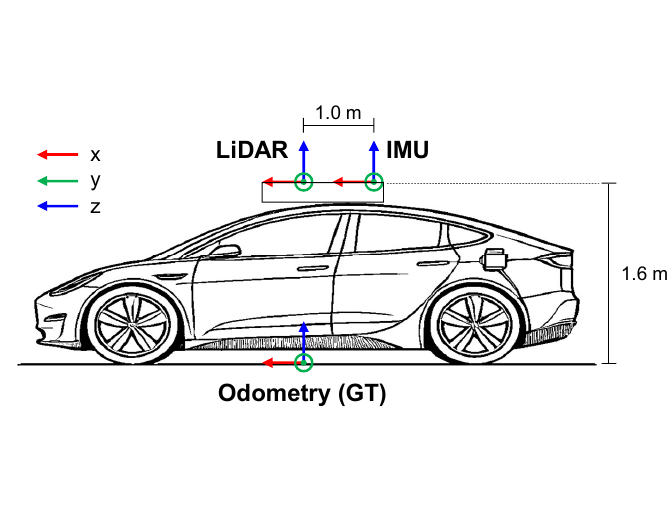}
    \caption{System Configuration}
    \label{fig:system_config}
\end{figure}

\subsection{System Configuration}
We configure the following virtual sensors in CARLA: a LiDAR and an IMU. Notably, we do not include a GNSS sensor because CARLA provides a built-in odometry message that already serves as a global ground-truth for vehicle poses. The simulation runs at 20Hz, with all sensor measurements synchronized at each frame.

\textbf{LiDAR.}
We employ a 32-channel LiDAR sensor, modeled after the \emph{Ouster OS1} design, mounted on the roof of a simulated vehicle. The vertical field of view is set to cover $\pm 22.5^{\circ}$, with a maximum range of 120m. Because the sensor data is generated within a simulation, each full scan is inherently \textit{deskewed} (i.e., free of motion-induced distortions). 

\textbf{IMU.}
We employ an IMU that provides \textit{ideal} ground-truth inertial data—namely, zero bias and noise.

\begin{figure}[t]
    \centering
    \includegraphics[width=0.5\textwidth]{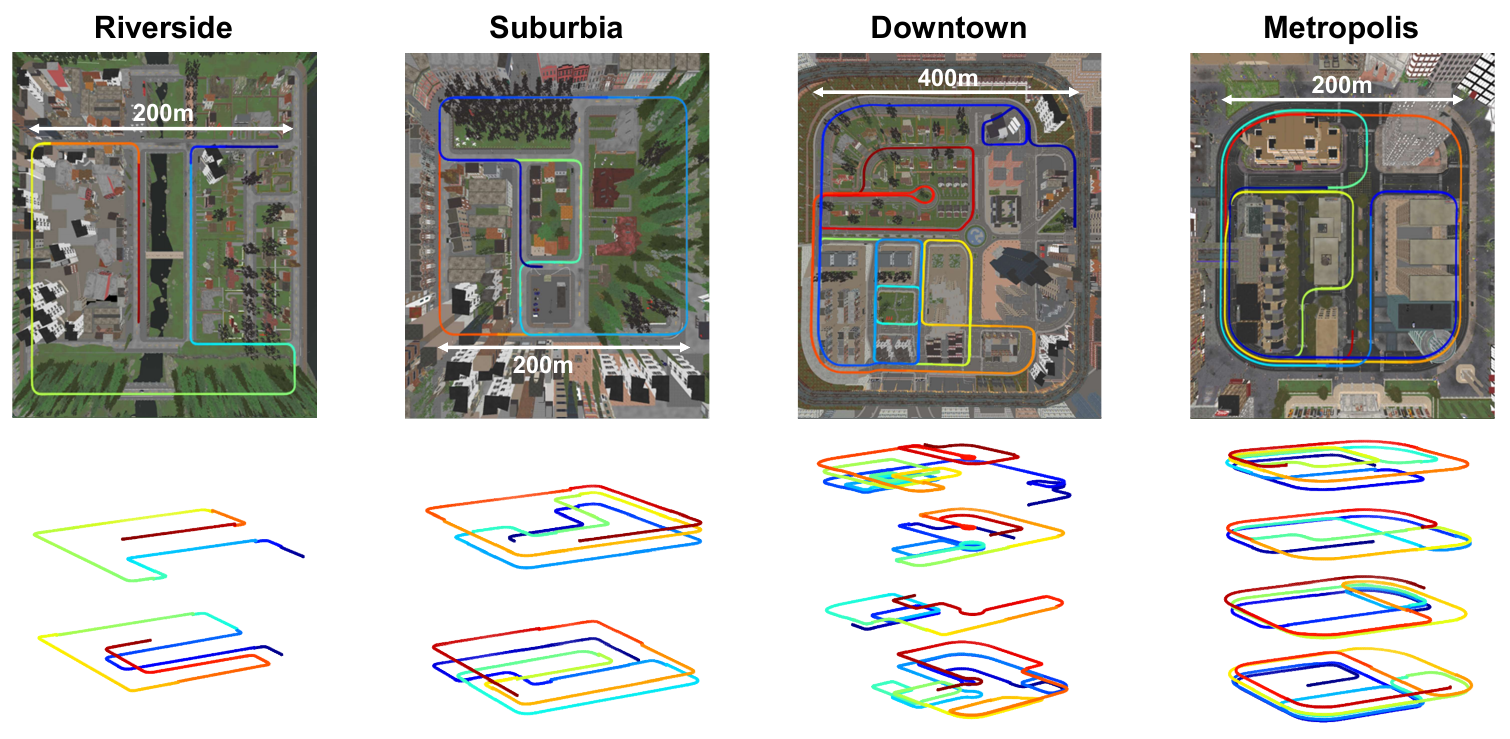}
    \caption{The upper displays aerial views with trajectories overlaid for \texttt{Riverside 02}, \texttt{Suburbia 02}, \texttt{Downtown 04}, and \texttt{Metropolis 04}. The below shows all the routes on each map, with red being the starting point and blue being the ending point.}
    \label{fig:data_generation_traj}
    \vspace{-6mm}
\end{figure}

\subsection{Dataset Generation} To create environments that challenge long-term PR, we introduce substantial structural modifications in CARLA. Starting with base maps \texttt{Riverside, Suburbia, Downtown}, and \texttt{Metropolis}, we manually delete key buildings, trees, and landmarks using the Unreal Engine Editor across multiple sequences. As illustrated in \figref{fig:data_generation_traj}, each modified sequence is then traversed by a scripted autonomous driving agent following partially overlapping, yet distinct routes, capturing structural changes from various viewpoints and ensuring meaningful PR experiments.

\subsection{Sequence Description} Our dataset comprises 4 maps and 12 sequences in total: \texttt{Downtown} and \texttt{Metropolis} each include 4 sequences (\texttt{01}–\texttt{04}), while \texttt{Riverside} and \texttt{Suburbia} each contain 2 sequences (\texttt{01}–\texttt{02}). Each sequence reflects a distinct stage of structural changes, where lower-numbered sequences represent maps with fewer structures, and higher-numbered sequences (e.g., \texttt{04} or \texttt{02}) represent the original, fully built environment. Thus, moving from lower to higher numbers simulates construction, and from higher to lower numbers simulates demolition as shown in \figref{fig:main}.

\section{Temporal Change Ratio}
\label{sec:method2}

\begin{figure}[!t]
    \centering
    \includegraphics[width=\columnwidth]{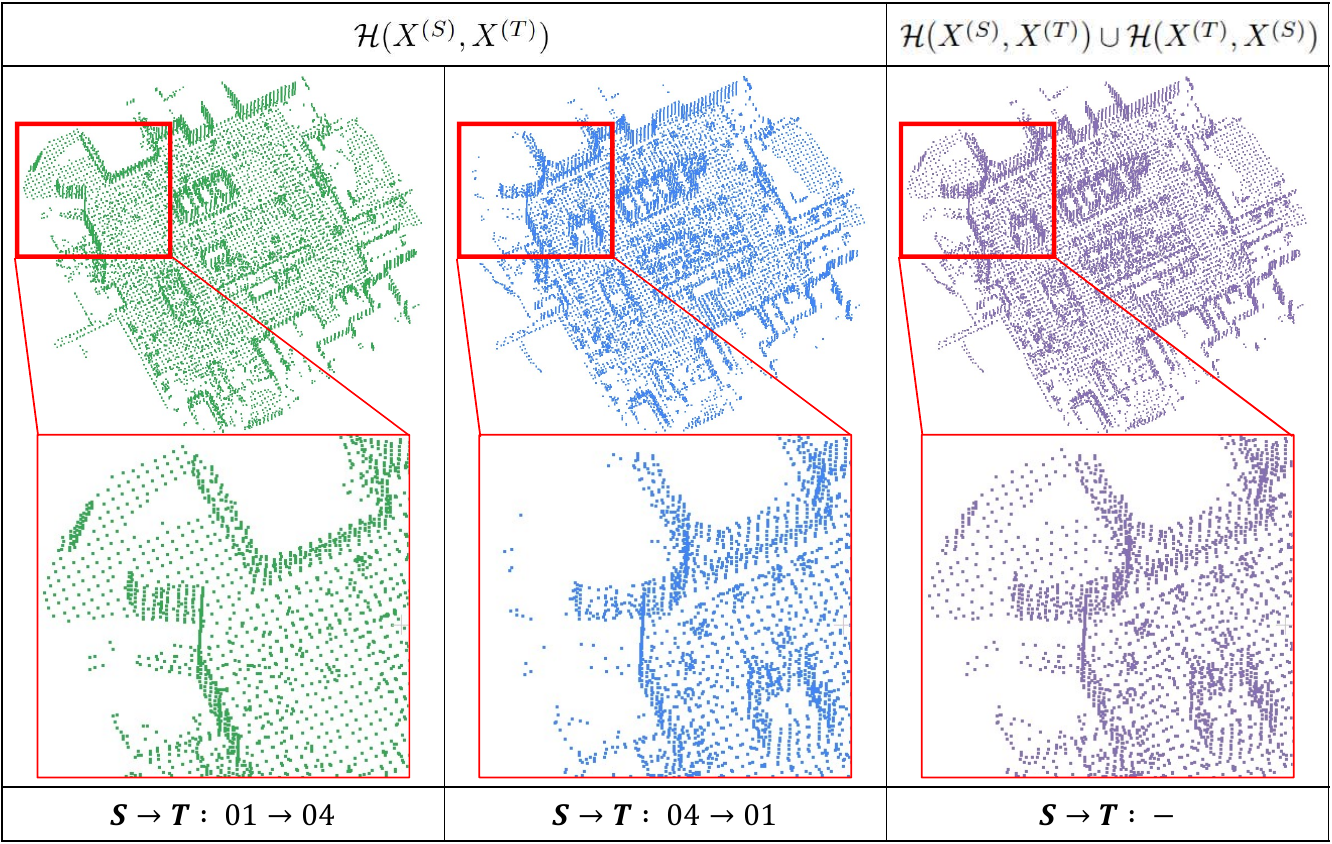}
    \caption{\textbf{Visualization of the set \( \mathcal{H} \) for} \texttt{Metropolis}{.} The left two columns illustrate how \(\mathcal{H}\) changes under the source–target ordering, causing variations in TCR values. To ensure consistency regardless of the ordering, we introduce the union of both sets, as shown in the right column, which remains invariant to the ordering.}
    \vspace{-3mm}
    \label{fig:TCR_visualization}
\end{figure}

We propose the TCR$_{\text{sym}}$ (symmetric Temporal Change Ratio) metric to quantify large-scale structural changes between sequences. Unlike the original TCR from the \ac{NSS} benchmark \cite{sun2025standsstillspatiotemporalbenchmark}, which is sensitive to the source-target ordering, TCR$_{\text{sym}}$ ensures consistent measurements of structural changes irrespective of the ordering. As shown in \figref{fig:TCR_visualization}, the original TCR can vary depending on the source-target direction, while the symmetric version eliminates this dependency. The TCR$_{\text{sym}}$ is defined as:
\begin{equation}
 \label{eq:tcr_sym} 
 \mathrm{TCR}_{\text{sym}} = 1 - \frac{\left|\mathcal{O}(X^{(S)},X^{(T)};\tau) \cup \mathcal{O}(X^{(T)},X^{(S)};\tau)\right|} {\left|\mathcal{H}(X^{(S)},X^{(T)}) \cup \mathcal{H}(X^{(T)},X^{(S)})\right|},
\end{equation}
where the unchanged point set \( \mathcal{O} \) and the convex hull set \( \mathcal{H} \) are defined as follows:
\begin{equation}
\label{eq:overlap}
    \mathcal{O}(X^{(S)}, X^{(T)}; \tau) 
    = \Bigl\{\,x \in X^{(S)} \,\Bigl\vert\, \mathrm{NN}(x, X^{(T)}) \le \tau \Bigr\},
\end{equation}
\begin{align}
\label{eq:convexhull}
    \mathcal{H}(X^{(S)}, X^{(T)}) &= \Bigl\{\,x \in X^{(S)} \,\Bigl\vert\, \mathrm{ConvexHull}\bigl(X^{(T)}\bigr) \notag \\
    &\quad \;\;\;\; = \mathrm{ConvexHull}\bigl(X^{(T)} \cup \{x\}\bigr)\Bigr\}.
\end{align}

\begin{table}[!t]
\caption{\centering Temporal Change Ratio comparison}
\label{tab:TCR}
\begin{adjustbox}{width=\columnwidth}
\begin{tabular}{l|cccc|c}
\toprule
\textbf{Map}       & \texttt{Riverside}                         & \texttt{Suburbia}                         & \texttt{Downtown}                         & \texttt{Metropolis}                         & \texttt{KAIST*}                \\ \midrule
\textbf{Sequences} & {\( 01 \leftrightarrow 02 \)} & {\( 01 \leftrightarrow 02 \)} & {\( 01 \leftrightarrow 04 \)} & {\( 01 \leftrightarrow 04 \)} & {\( 01 \leftrightarrow 06 \)} \\ \midrule
TCR$_{\text{sym}}$       & 0.0986                         & 0.1840                         & 0.1120                         & 0.2920                         & 0.1819                         \\ \bottomrule
\end{tabular}
\end{adjustbox}
\vspace{-5mm}
\end{table}
\label{fig:TCR_tab}

A higher TCR$_{\text{sym}}$ indicates more structural changes, such as significant building construction or demolition, whereas a lower TCR$_{\text{sym}}$ reflects smaller-scale modifications. To focus primarily on structural rather than transient changes (e.g., vehicles, pedestrians), we voxelize the point clouds at a resolution of 5m and set the distance threshold \(\tau = 4.5\)m. This ensures TCR$_{\text{sym}}$ predominantly captures large-scale structural changes rather than transient objects.

Table~\ref{tab:TCR} summarizes the TCR$_{\text{sym}}$ values observed between sequences in our dataset alongside a real-world comparison. The \texttt{KAIST} entry represents the structural changes occurring over more than four years (from MulRan's \texttt{KAIST 01} to HeLiPR's \texttt{KAIST 06} sequences), aligned using the LT-SLAM module from LT-mapper \cite{kim2022lt}. Notably, our dataset exhibits TCR$_{\text{sym}}$ values that meet or even surpass the changes observed in the KAIST sequences. This highlights the capability of our dataset to effectively emulate extensive long-term urban transformations, offering a more rigorous benchmark for long-term PR research.
\section{Benchmark Results with Our Dataset}
\label{sec:method}

In this section, we evaluate representative LiDAR-based PR methods—Scan Context \cite{kim2021scan}, SOLiD \cite{kim2024narrowing}, RING++ \cite{xu2023ring++}, and BTC \cite{yuan2024btc}—on our \ac{CNS} dataset. We compare these baselines across varying degrees of structural changes, using standard metrics such as the Precision–Recall curve, Area Under the Curve (AUC), Recall at N (\(R@N\)) and the maximum F1 score along with TCR$_{\text{sym}}$.

\begin{table*}[]
\centering
\caption{Quantitative results for long-term PR}
\label{tab:PR}
\resizebox{\textwidth}{!}{%
\begin{tabular}{@{}ccccccccccccccc@{}}
\toprule
\multirow{3}{*}{Map}
  & \multirow{3}{*}{\(\text{DB} \rightarrow \textit{query}\)}
  & \multirow{3}{*}{\(\mathrm{TCR}_{\text{sym}}\)}
  & \multicolumn{12}{c}{Method} \\      
\cmidrule(l){4-15}                                      
      & & &
        \multicolumn{3}{c}{Scan Context++~\cite{kim2021scan}} &
        \multicolumn{3}{c}{SOLiD~\cite{kim2024narrowing}} &
        \multicolumn{3}{c}{RING++~\cite{xu2023ring++}} &
        \multicolumn{3}{c}{BTC~\cite{yuan2024btc}} \\

      & & & AUC & R@1 & F1 score & AUC & R@1 & F1 score & AUC & R@1 & F1 score & AUC & R@1 & F1 score \\ \midrule
\multirow{2}{*}{ \texttt{Riverside}}
  & \multicolumn{1}{c|}{\(01 \rightarrow 02\)} & \multirow{2}{*}{0.0986}
  & \multicolumn{1}{|c}{0.1423} & 0.1271 & 0.2222 & 0.0703 & 0.0678 & 0.1119
  & \textbf{0.8856} & \textbf{0.7664} & \textbf{0.8678}
  & \underline{0.4126} & \underline{0.4538} & \underline{0.6178} \\
  & \multicolumn{1}{c|}{\(02 \rightarrow 01\)} &
  & \multicolumn{1}{|c}{0.1268} & 0.1695 & 0.2032 & 0.1199 & 0.1356 & 0.1827
  & \textbf{0.8291} & \textbf{0.9626} & \textbf{0.8653}
  & \underline{0.2702} & \underline{0.4091} & \underline{0.5243} \\ \midrule
\multirow{2}{*}{ \texttt{Suburbia}}
& \multicolumn{1}{c|}{\(01 \rightarrow 02\)} & \multirow{2}{*}{0.1840}
& \multicolumn{1}{|c}{0.4480} & 0.3732 & 0.5146 & 0.2066 & 0.1901 & 0.3094
& \textbf{0.8974} & \textbf{0.7192} & \textbf{0.8367}
& \underline{0.6359} & \underline{0.6414} & \underline{0.7706} \\
& \multicolumn{1}{c|}{\(02 \rightarrow 01\)} &
& \multicolumn{1}{|c}{0.4699} & 0.3285 & 0.4839 & 0.1442 & 0.1168 & 0.2000
& \textbf{0.8872} & \textbf{0.6667} & \textbf{0.8000}
& \underline{0.5479} & \underline{0.5521} & \underline{0.6928} \\ \midrule
\multirow{3}{*}{ \texttt{Downtown}}
& \multicolumn{1}{c|}{\(03 \rightarrow 04\)} & 0.0678
& \multicolumn{1}{|c}{0.3567} & 0.4560 & 0.5192 & \underline{0.5530} & \underline{0.5520} & \underline{0.5308}
& \textbf{0.9626} & \textbf{0.9816} & \textbf{0.9907}
& 0.2907 & 0.3828 & 0.4923 \\
& \multicolumn{1}{c|}{\(02 \rightarrow 04\)} & 0.1120
& \multicolumn{1}{|c}{\underline{0.4734}} & \underline{0.5674} & \underline{0.5613} & 0.3007 & 0.3333 & 0.3800
& \textbf{0.9422} & \textbf{0.8945} & \textbf{0.9443}
& 0.2554 & 0.3147 & 0.4562 \\
& \multicolumn{1}{c|}{\(01 \rightarrow 04\)} & 0.1674
& \multicolumn{1}{|c}{0.2204} & 0.2303 & 0.2833 & 0.0796 & 0.1292 & 0.1402
& \textbf{0.8131} & \textbf{0.7038} & \textbf{0.8262}
& \underline{0.3450} & \underline{0.4532} & \underline{0.5986} \\ \midrule
\multirow{3}{*}{ \texttt{Metropolis}}
& \multicolumn{1}{c|}{\(03 \rightarrow 04\)} & 0.2528
& \multicolumn{1}{|c}{\underline{0.6566}} & 0.5185 & 0.6693 & 0.4220 & 0.2963 & 0.4495
& \textbf{0.9550} & \textbf{0.8721} & \textbf{0.9317}
& 0.5456 & \underline{0.5605} & \underline{0.7092} \\
& \multicolumn{1}{c|}{\(02 \rightarrow 04\)} & 0.2552
& \multicolumn{1}{|c}{0.3566} & 0.2680 & 0.4039 & 0.3015 & 0.2288 & 0.3431
& \textbf{0.8341} & \textbf{0.6807} & \textbf{0.8100}
& \underline{0.5658} & \underline{0.5877} & \underline{0.7037} \\
& \multicolumn{1}{c|}{\(01 \rightarrow 04\)} & 0.2920
& \multicolumn{1}{|c}{0.4442} & 0.2993 & 0.4623 & 0.2582 & 0.2449 & 0.3512
& \textbf{0.7228} & \textbf{0.5837} & \textbf{0.7371}
& \underline{0.5164} & \underline{0.5362} & \underline{0.6756} \\ \bottomrule
\end{tabular}}
\vspace{-4mm}
\end{table*}


For the PR task, we preprocess our dataset as follows. Query sequences are sampled at intervals of 10 meters along their trajectories, while database sequences are sampled at 5 meters intervals. Given a query scan, candidate matches are retrieved from the database based on their spatial similarity. A candidate is considered a \textit{true positive} if its Euclidean distance to the query location is within 7.5 meters. Additionally, to ensure consistency across methods, each raw LiDAR point cloud is uniformly limited to a range of [-100m, 100m]. For evaluating BTC specifically, we merge 10 sequential scans into one submap as described in the original paper.

\begin{figure}[t]
    \centering
    \includegraphics[width=\linewidth]
    {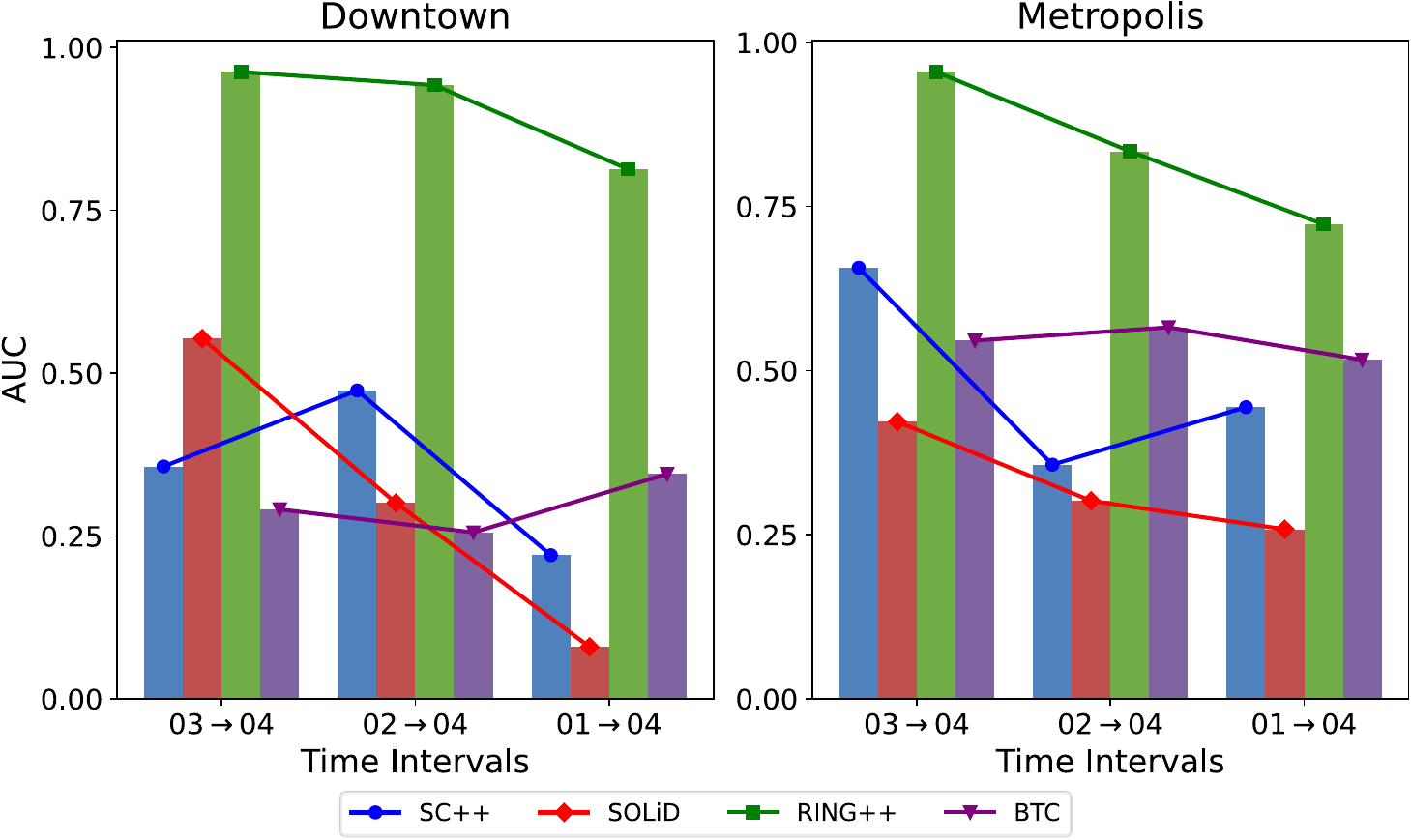}
    \caption{AUC for each baseline as time intervals increase}
    \label{fig:AUC_TCR}
    \vspace{-7mm}
\end{figure}

\begin{figure}[t]
    \centering
    \includegraphics[width=\linewidth]
    {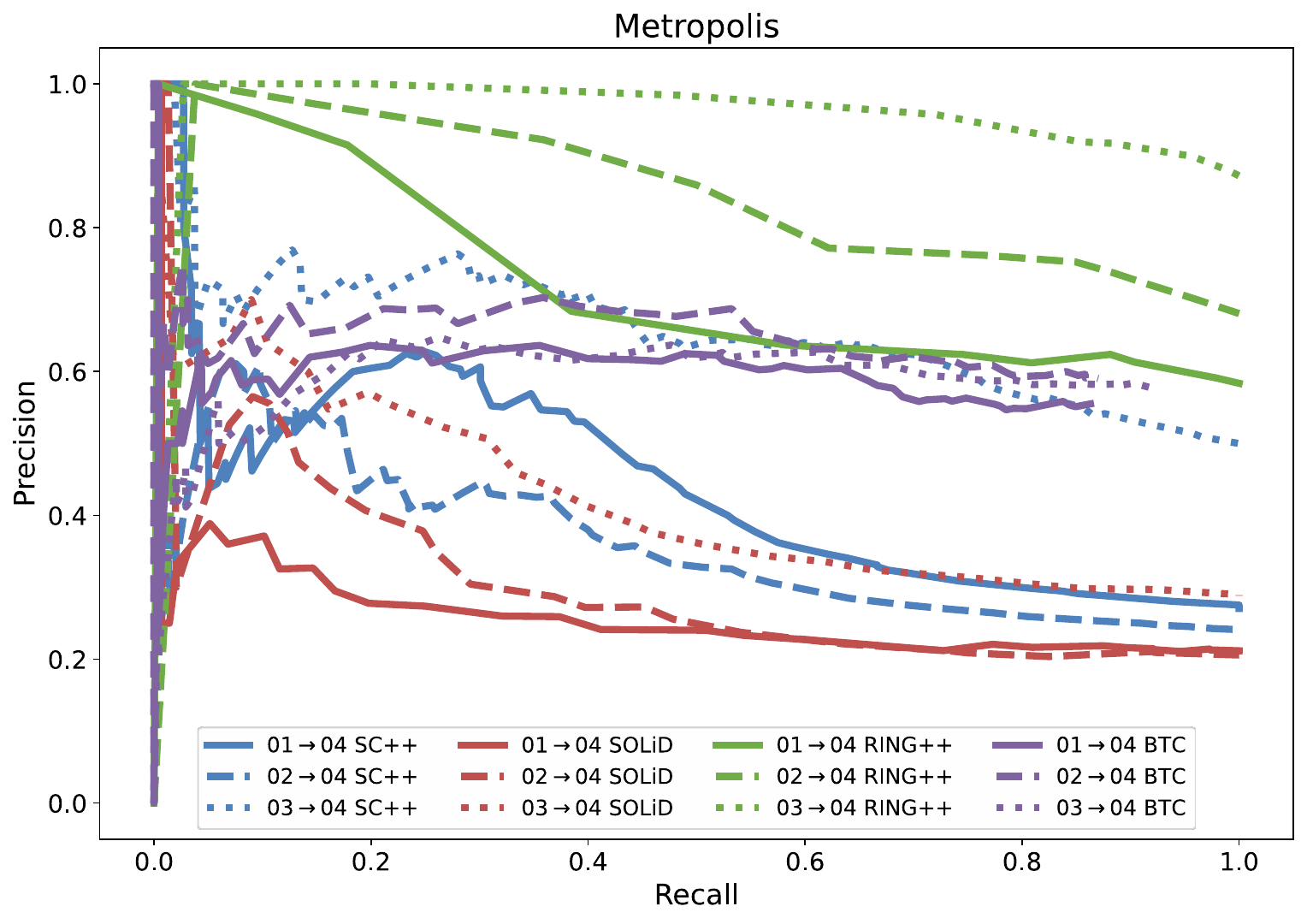}
    \caption{Precision–Recall curves of each baseline for \texttt {Metropolis}}
    \label{fig:PRcurve_Downtown_&_Metropolis}
    \vspace{-1.5mm}
\end{figure}

Table~\ref{tab:PR} provides the AUC, \(R@N\) and the F1 score for each method on different inter-session pairs of our dataset. Bold and underlined values indicate the best and second-best performance, respectively. As observed in the Table~\ref{tab:PR} and Fig~\ref{fig:AUC_TCR}, PR performance decreases as temporal change increases. This indicates that, as the structural changes in the environment become more significant, PR algorithms generally face increasing difficulty in maintaining high accuracy.

However, BTC exhibits relatively consistent performance regardless of the increase in TCR$_{\text{sym}}$. This can be attributed to its effective local descriptor construction from the submaps and the robust verification logic applied during candidate matching. The recall values of BTC increase steadily, while maintaining relatively high precision, as seen in the Fig~\ref{fig:PRcurve_Downtown_&_Metropolis}. Despite this, the absolute performance of BTC remains lower compared to other methods, which suggests limitations in its ability to handle significant transformations effectively.

In contrast, RING++ stands out with superior performance compared to the other methods. This can be explained by its use of Bird’s Eye View (BEV) images, which incorporate contour information from roads and surrounding distances. This strategy appears to be highly effective for long-term PR tasks, as it allows the method to capture broader contextual information. However, RING++ shows a clear performance drop as TCR$_{\text{sym}}$ increases, revealing a limitation in handling extreme environmental changes.

In summary, while BTC and RING++ show resilience to structural changes, they still exhibit significant performance degradation as TCR$_{\text{sym}}$ rises. This emphasizes the need for further improvements in long-term PR algorithms, particularly in adapting to large-scale transformations.

\section{Conclusion}
\label{sec:conculsion}
The \ac{CNS} dataset pushes the boundaries of long-term PR by incorporating large-scale structural transformations in both urban and suburban settings, all within a reproducible simulation environment. Our evaluation of multiple LiDAR-based PR methods shows that, although certain algorithms maintain resilience under moderate changes, they experience pronounced performance drops as TCR$_{\text{sym}}$ increases. This outcome underscores the need for novel descriptors and strategies that can accommodate massive redevelopment scenarios. We hope that by providing reproducible evaluations and TCR$_{\text{sym}}$-based analysis, our \ac{CNS} dataset can catalyze further advances in long-term PR for real-world construction and demolition tasks.


\balance
\small
\bibliographystyle{IEEEtranN} 
\bibliography{string-short,references}

\end{document}